\documentclass[10pt,twocolumn,letterpaper]{article}

\usepackage[pagenumbers]{cvpr} % To force page numbers, e.g. for an arXiv version
\usepackage[numbers]{natbib}
\usepackage[most]{tcolorbox}

\newtcolorbox{disclaimerbox}{
    colback=gray!10,     % background color
    colframe=gray!40,    % frame color
    boxrule=0.5pt,       % border thickness
    arc=4pt,             % rounded corners
    auto outer arc,
    boxsep=5pt,
    left=6pt,
    right=6pt,
    top=4pt,
    bottom=4pt,
    enhanced jigsaw
}

\definecolor{iccvblue}{rgb}{0.21,0.49,0.74}
\usepackage[pagebackref,breaklinks,colorlinks,allcolors=iccvblue]{hyperref}

\def\paperID{*****} % *** Enter the Paper ID here
\def\confName{CVPR}
\def\confYear{2026}

\title{ELoG-GS: Dual-Branch Gaussian Splatting with Luminance-Guided Enhancement for Extreme Low-light 3D Reconstruction}

\author{
Yuhao Liu$^{1}$ \quad
Dingju Wang$^{1}$ \quad
Ziyang Zheng$^{1}$\\ \\
$^{1}$\,Shanghai Jiao Tong University \\ \\
\{liuyuhao, wangdingju, zhengziyang\}@sjtu.edu.cn
}

\begin{document}
\maketitle

\begin{abstract}
% A brief summary of the problem, i.e., challenging light conditions in 3D Reconstruction and real-world applications. Short description of the proposed method. The proposed method was presented in the NTIRE 3D Restoration and Reconstruction (3DRR) Challenge, and outperformed the baseline methods by a large margin. The code is available at \url{}.
This paper presents our approach to the NTIRE 2026 3D Restoration and Reconstruction Challenge (Track 1), which focuses on reconstructing high-quality 3D representations from degraded multi-view inputs. The challenge involves recovering geometrically consistent and photorealistic 3D scenes in extreme low-light environments. To address this task, we propose \textit{Extreme Low-light Optimized Gaussian Splatting (ELoG-GS)}, a robust low-light 3D reconstruction pipeline that integrates learning-based point cloud initialization and luminance-guided color enhancement for stable and photorealistic Gaussian Splatting. Our method incorporates both geometry-aware initialization and photometric adaptation strategies to improve reconstruction fidelity under challenging conditions. Extensive experiments on the NTIRE Track 1 benchmark demonstrate that our approach significantly improves reconstruction quality over the baselines, achieving superior visual fidelity and geometric consistency. The proposed method provides a practical solution for robust 3D reconstruction in real-world degraded scenarios. In the final testing phase, our method achieved a PSNR of 18.6626 and an SSIM of 0.6855 on the official platform leaderboard. Code is available at: \url{https://github.com/lyh120/FSGS_EAPGS}.
\end{abstract}

% ----------------------------- Optional -----------------------------
\begin{disclaimerbox}
Our method achieved a ranking of 9 out of 148 participants in Track 1 of the NTIRE 3DRR Challenge, as reported on the official competition website: \url{https://www.codabench.org/competitions/13854/}.
\end{disclaimerbox}
% --------------------------------------------------------------------

\section{Introduction}
\label{sec:intro}
    Recent restoration methods have made notable progress in handling real-world adverse degradations, including low-light enhancement, smoke and haze removal, denoising, and super-resolution~\cite{liu2026ntire,chang2026training,ge2026dual,chang2026beyond,ge2026clip}. More recently, these restoration techniques have also been incorporated into 3D Gaussian Splatting pipelines to improve novel view synthesis and reconstruction from degraded multi-view observations~\cite{zheng20263d,liu2026elog,fu2026smokegs,cao2026gensmoke,zhu2026naka,guo2026reliability,chen2026dehaze}.
    
    The proposed \textbf{ELoG-GS} is motivated by the severe degradation of traditional Structure-from-Motion (SfM)~\cite{schonberger2016structure} pipelines and vanilla 3D Gaussian Splatting (3DGS)~\cite{kerbl20233d} in near-zero illumination and extremely sparse-view (approximately 30 images) conditions. Under such conditions, standard feature matching fails due to pervasive sensor noise, while sparse constraints lead to geometric collapse and severe color drift. To overcome these challenges, we adopt an explicit ``restoration-then-reconstruction'' decoupling framework optimized for extreme low-light 3D synthesis.
    
    In the preprocessing stage, we utilize \textbf{Retinexformer}~\cite{cai2023retinexformer} (pre-trained on LOL\_v2\_real) as a zero-shot restoration engine to recover latent scene information from the darkness. For robust geometric initialization, we leverage \textbf{VGGT}~\cite{wang2025vggt}-based depth estimation and a voxelized fusion strategy to bypass unreliable SfM results. Our reconstruction engine features a hybrid dual-branch architecture integrating \textbf{FSGS}~\cite{zhu2024fsgs} and \textbf{EAP-GS}~\cite{dai2025eap}. The FSGS branch is customized with a random initialization strategy and regularized optimization to ensure global stability, while the EAP-GS branch utilizes camera-pose conversion and monocular depth priors to recover high-frequency textures and sharp boundaries. Finally, rendered outputs undergo a luminance-guided enhancement and histogram matching to align pixel distributions with physical scene statistics, effectively eliminating color artifacts and ensuring photorealistic fidelity. During this process, both branches are evaluated, and the better-performing one is selected.

\begin{figure*}[t]
    \centering
    \includegraphics[width=\textwidth]{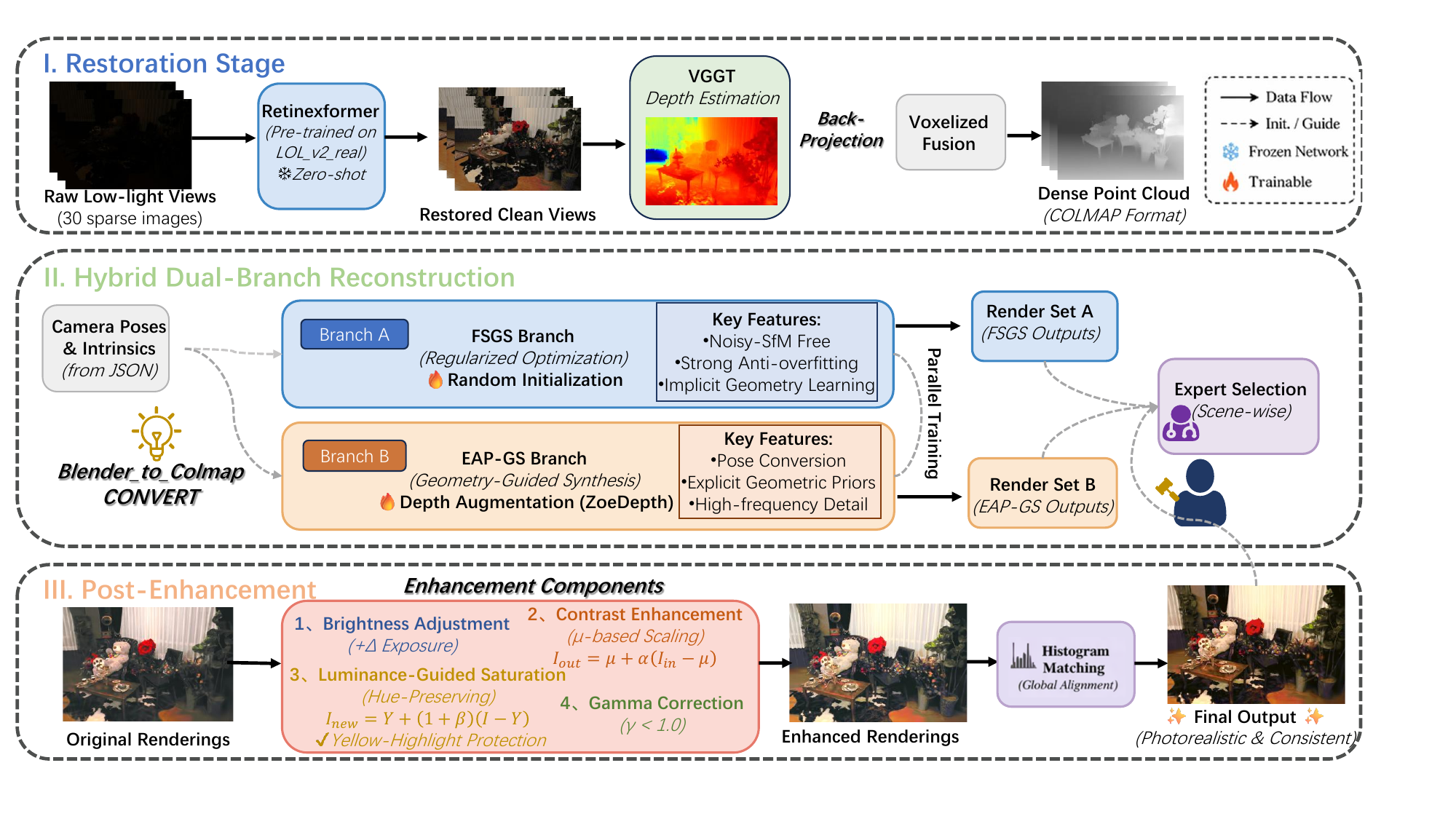}
    \caption{Overview of the ELoG-GS pipeline. \textbf{Stage I (Restoration)}: Raw low-light multi-view images are processed by a pre-trained Retinexformer (frozen) for zero-shot illumination recovery, while VGGT produces per-view depth maps that are back-projected and voxel-fused into a dense, COLMAP-compatible point cloud. \textbf{Stage II (Hybrid Dual-Branch Reconstruction)}: Branch A (FSGS) performs regularized optimization from random initialization without SfM dependency; Branch B (EAP-GS) leverages converted camera poses and ZoeDepth-based point cloud augmentation for geometry-guided synthesis. Both branches are trained in parallel, and an expert selection mechanism chooses the better reconstruction per scene. \textbf{Stage III (Post-Enhancement)}: Rendered outputs undergo brightness adjustment, mean-based contrast scaling, luminance-guided hue-preserving saturation enhancement, and gamma correction, followed by histogram matching for globally consistent color reproduction.}
    \label{fig:arch}
\end{figure*}

    The main contributions of our work include: an optimized decoupling framework for extreme low-light scenarios, a hybrid dual-branch architecture for sparse-view 3DGS that balances geometric regularity with textural detail, and a robust post-processing pipeline for consistent color reproduction. The overall architecture of ELoG-GS, featuring restoration, hybrid dual-branch reconstruction, and post-enhancement stages, is illustrated in Fig.~\ref{fig:arch}.

\section{Related Work}
\label{sec:related_work}

% Literature review, including the references outlined in main.bib

\noindent \textbf{Low-light Image Enhancement and Restoration:} Research in this domain fundamentally aims to enhance visibility, suppress noise, and recover structural cues from severely underexposed photographs. Substantial progress has been achieved in 2D image enhancement, with representative works including Retinex-based decomposition models \cite{wei2018deep,zhang2019kindling, cai2023retinexformer}, diffusion-driven restoration techniques \cite{jiang2023low}, and data-driven methods tailored for precise illumination correction. These approaches consistently demonstrate high fidelity in exposure and color recovery, yielding visually pleasing and well-illuminated results. Consequently, they  can serve as a highly effective and stable pre-processing foundation for downstream perception and 3D reconstruction tasks, successfully mitigating the impacts of severe signal degradation in multi-view observations.

% \subsection{Challenges in Multi-view Reconstruction Under Extreme Low-light}
% Traditional structure-from-motion (SfM) and multi-view stereo (MVS) techniques struggle significantly when image quality deteriorates due to ultra-low illumination. Feature detection and matching become unreliable, leading to fragmented tracks and sparse point clouds. In such conditions, conventional pipelines may generate unstable camera poses or fail entirely, making robust geometric initialization crucial. Comparisons against well-lit datasets highlight the dramatic drop in feature density and reconstruction completeness, motivating the design of learning-based geometry priors.

% \noindent \textbf{3D Gaussian Splatting (3DGS):} 3DGS has emerged as an efficient alternative to volumetric radiance fields by enabling fast rendering and real-time optimization. However, its susceptibility to overfitting introduces artifacts such as floating points and geometry collapse when input views are sparse or degraded. Improved variants propose strong regularization, depth-aware constraints, or external point cloud guidance to enhance robustness. These refinements demonstrate the importance of reliable initialization and photometric consistency when adapting 3DGS to challenging conditions.
\noindent \textbf{3D Gaussian Splatting (3DGS):} While 3DGS has revolutionized novel view synthesis with real-time rendering and efficient optimization, it is highly susceptible to overfitting—often introducing floating artifacts and geometry collapse under sparse-view (few-shot) or degraded conditions. To tackle this, recent advancements explicitly target robust geometric initialization and structure regularization. Notably, EAP-GS \cite{dai2025eap} mitigates scene degradation by enriching sparse SfM priors via attentional point cloud augmentation, emphasizing the critical role of adequate initialization. Concurrently, FSGS \cite{zhu2024fsgs} addresses extreme sparsity through proximity-guided Gaussian unpooling coupled with depth-aware pseudo-view regularization. These tailored refinements underscore that reliable point initialization and consistent geometric constraints are indispensable for robust 3DGS optimization in challenging few-shot scenarios. Beyond geometric initialization and regularization, recent works have also explored improving the representation capacity of splatting primitives themselves. For instance, 3DGabSplat \cite{zhou20253dgabsplat} replaces Gaussian kernels with 3D Gabor-based primitives to enhance the capture of high-frequency details, achieving superior rendering quality with fewer primitives.

\noindent \textbf{Learning-Based Geometric Initialization and Feed-forward Depth Estimation:}
To address the degradation of classical SfM under low-light or weak-texture conditions, recent advances leverage feed-forward neural networks as geometric priors. Methods such as VGGT~\cite{wang2025vggt}, DUSt3R~\cite{wang2024dust3r}, and Mast3R~\cite{leroy2024grounding} directly predict dense geometry (e.g., depth, point maps, or cross-view correspondences) from single or sparse multi-view inputs, enabling implicit cross-view alignment without explicit feature matching. These predictions can be further consolidated via volumetric or point-based fusion to improve global consistency, and serve as robust initialization for subsequent 3D representations (e.g., 3DGS). Under sparse views or degraded imaging conditions, such learned priors significantly enhance reconstruction robustness and completeness.
% \subsection{Photometric and Illumination Modeling in 3D Reconstruction}
% Illumination inconsistencies are a major obstacle for reconstructing scenes captured under low-light or view-dependent brightness variations. Recent methods introduce luminance-guided optimization, radiance stabilization, or structural transformations that disentangle lighting changes across viewpoints. These techniques improve color stability and mitigate artifacts caused by uncontrolled illumination, enabling more faithful reconstruction when raw images present inconsistent exposure or severe noise.

% \subsection{Disentangled Appearance-Geometry Representations}
% Methods that separate geometry from appearance variations—such as transient-aware or environmental illumination–aware radiance field models—have shown effectiveness in handling dynamic lighting conditions. By explicitly modeling transient components or decomposing appearance into stable and variable factors, these approaches improve robustness in scenes with view-dependent or temporally varying effects. Their principles provide conceptual grounding for designing systems tailored to low-light multi-view reconstruction.

\section{Methods}
\label{subsec:methods}

% \subsection{Extreme Low-light Optimized Gaussian Splatting }

We propose Extreme Low-light Optimized Gaussian Splatting (ELoG-GS), a robust 3D reconstruction pipeline tailored for extremely low-light and sparse-view (approximately 30 images) multi-view inputs. ELoG-GS follows an explicit \textit{restoration-then-reconstruction} decoupling framework to address severe sensor noise, feature matching failure, geometric collapse, and color drift in conventional 3DGS pipelines. The pipeline comprises three core stages: preprocessing (point-cloud initialization and low-light image restoration), hybrid dual-branch 3D reconstruction, and luminance-guided post-processing.

\subsection{Preprocessing}
\subsubsection{Learning-based Point Cloud Initialization}
\label{sec:vggt-colmap-init}
Traditional SfM via COLMAP fails under extremely low-light or few-shot conditions, producing excessively sparse point clouds (fewer than 300 points) due to unreliable feature detection and matching. %\cite{schonberger2016structure}. 
To resolve this, we employ VGGT %\cite{wang2025vggt} 
for depth estimation from degraded multi-view images. Based on predicted depth maps and camera parameters, we back-project pixels to 3D space and apply a voxelized fusion strategy: 3D points within the same voxel are averaged, and only voxels with sufficient observations are retained. This strategy suppresses noise, enforces multi-view consistency, and generates a dense, geometrically reliable point cloud converted to COLMAP-compatible format for downstream 3DGS optimization.
%\cite{kerbl20233d}.

\subsubsection{Zero-shot Low-light Image Restoration}
We abandon latent decoupling methods and diffusion-based models due to instability and artifacts under extreme photon starvation. Instead, we adopt Retinexformer
%\cite{cai2023retinexformer} 
as a zero-shot restoration engine. It effectively recovers latent scene information from near-zero illumination, suppresses sensor noise, and preserves color fidelity, providing clean inputs for subsequent 3D reconstruction.

\subsection{Hybrid Dual-branch 3D Reconstruction}
To mitigate overfitting, geometric collapse, and color drift in sparse-view vanilla 3DGS%\cite{kerbl20233d}
, we design a dual-branch architecture integrating Few-Shot Gaussian Splatting (FSGS)
%\cite{zhu2024fsgs} 
and Efficient Augmentation of Pointcloud for 3DGS (EAP-GS), with a hybrid selection strategy to choose the optimal branch per scene.

\subsubsection{Branch 1: Regularized FSGS with Random Initialization}
We modify FSGS to remove SfM dependency and use random initialization of 3D Gaussians within the scene bounding volume. Tuned hyperparameters for optimization (position learning rates, densification gradients) enable stable convergence via a regularized photometric loss. This branch delivers smooth, artifact-free global geometry and robust background reconstruction.

\subsubsection{Branch 2: Geometry-guided EAP-GS}
We convert camera parameters (extrinsics, intrinsics) from JSON format to COLMAP-compatible poses. EAP-GS leverages monocular depth priors (ZoeDepth)\cite{bhat2023zoedepth} to densify and augment the initial point cloud, guiding Gaussian splitting and cloning. This branch excels at recovering high-frequency textures, sharp edges, and complex foreground structures.

\subsubsection{Hybrid Selection Strategy}
After parallel training of both branches, we render novel views and apply a scene-specific selection strategy to select the optimal reconstruction. In our current pipeline, this branch selection relies on human-in-the-loop expert evaluation to maximize visual fidelity and geometric consistency.

\subsection{Luminance-guided Post-processing}
We apply a lightweight post-processing pipeline to rendered outputs for photorealistic color reproduction:
\begin{enumerate}
    \item Global brightness shift and gamma correction to compensate for underexposure and recover dark-region details.
    \item Contrast enhancement by scaling intensity deviation from the mean to improve clarity while maintaining luminance consistency.
    \item Luminance-based saturation adjustment to amplify chrominance residuals without hue shifts or color artifacts, especially in highlight regions.
\end{enumerate}
Histogram matching is further used to align pixel distributions with physical scene statistics, ensuring stable and consistent color reproduction across views.

\section{Experiments and Results}
\label{subsec:experiment}

   \begin{figure*}[htbp]
        \centering
        \includegraphics[width=\textwidth]{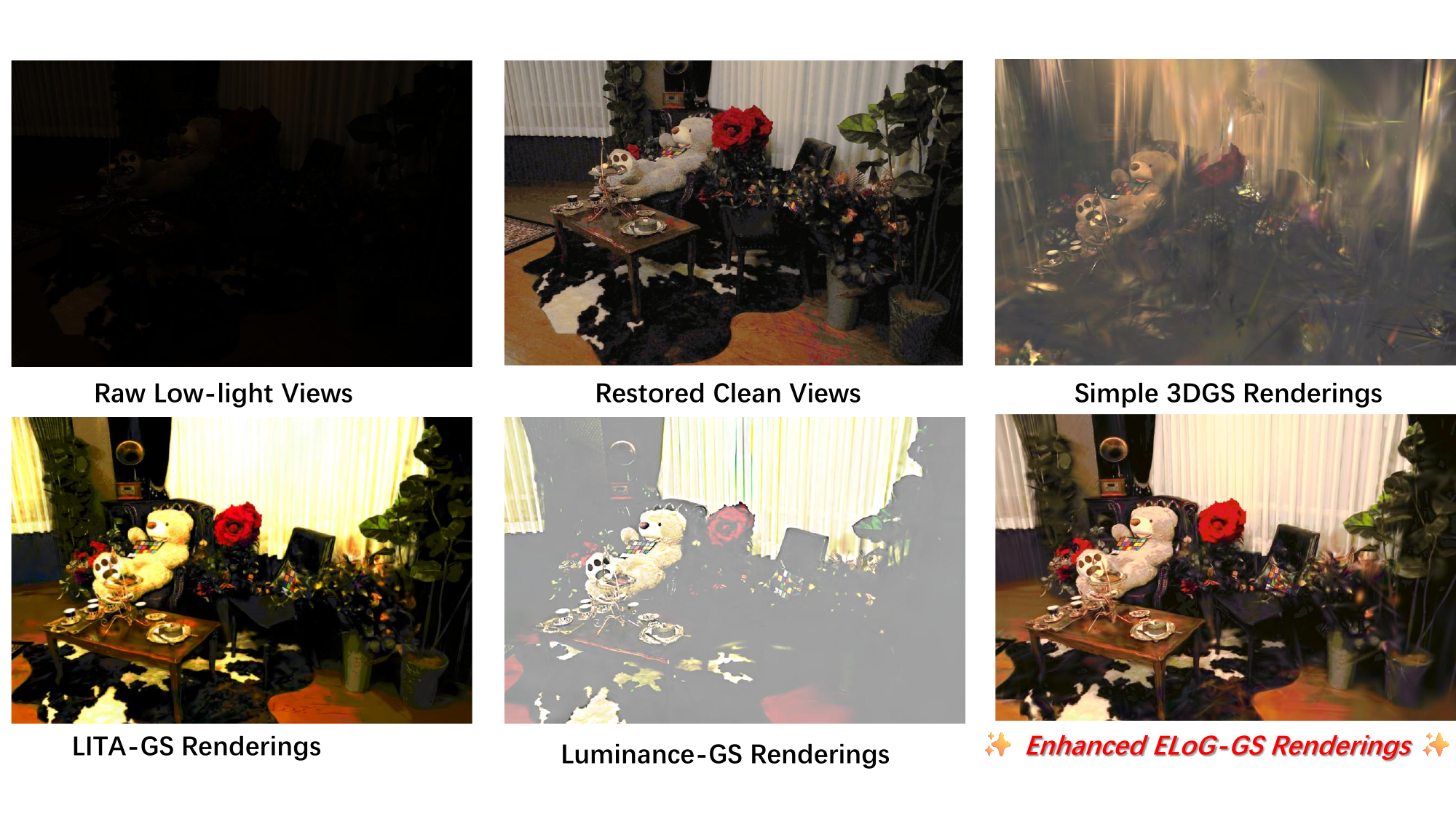}
        \caption{Qualitative comparison of different pipelines under extreme low-light conditions. From left to right: raw low-light inputs, restored images, vanilla 3DGS, LITA-GS, Luminance-GS, and enhanced ELoG-GS. }
        \label{fig:com}
    \end{figure*}
%\cite{zhou2025lita} \cite{Cui_2025_CVPR}
    
\noindent \textbf{Datasets.} We  train and evaluate ELoG-GS on the NTIRE 2026 challenge datasets \cite{liu2025realx3d}, which provide only a highly sparse set of views. 
% (approximately 30 images per scene).
In the final testing phase, the NTIRE 2026 challenge dataset comprises seven scenarios, each containing approximately 30 training scenes and six test scenes. All images are provided at their original resolution, which is close to 2K.

\noindent \textbf{Implementation Details.} ELoG-GS integrates two pipelines, namely FSGS and  EAP-GS, which are implemented in PyTorch and employ a CUDA-based differentiable Gaussian rasterizer. For the FSGS pipeline, instead of relying on COLMAP SfM, we initialize the 3D scene with a randomly generated sparse point cloud consisting of 100{,}000 points. All models are trained at the original image resolution for 30{,}000 iterations, using an optimizer with an initial learning rate of $1.6 \times 10^{-4}$ for Gaussian positions. This positional learning rate is decayed to $1.6 \times 10^{-6}$ over the course of optimization. For the EAP-GS pipeline, we initialize the reconstruction using the VGGT-generated COLMAP-compatible point cloud described in Section~\ref{sec:vggt-colmap-init}. All models are trained at the original image resolution for 5{,}000 iterations, using an optimizer with an initial learning rate of $1.6 \times 10^{-4}$ for Gaussian positions. This positional learning rate is decayed to $1.6 \times 10^{-6}$ over the course of optimization. We conduct our experiments on a workstation equipped with an NVIDIA RTX 4090 GPU.

\noindent \textbf{Results.} We compare our method against two state-of-the-art baselines released by the official source, namely LITA-GS ~\cite{zhou2025lita} and Luminance-GS ~\cite{Cui_2025_CVPR}. Evaluations are conducted on the competition datasets using PSNR ~\cite{HuynhThu2008ScopeOV} and SSIM ~\cite{1284395} as quantitative metrics. LITA-GS obtains 15.63 PSNR and 0.542 SSIM, while Luminance-GS achieves 10.89 PSNR and 0.531 SSIM. In comparison, our ELoG-GS achieves 18.66 PSNR and 0.685 SSIM. As shown in Fig.~\ref{fig:com}, the results demonstrate that our method outperforms the other SOTA methods. Notably, our method achieves better visual fidelity and geometric consistency, indicating that it preserves image information much more effectively.

\section{Conclusion}
\label{sec:conclusion}

This paper presents Extreme Low-light Optimized Gaussian Splatting (ELoG-GS), a robust 3D reconstruction pipeline for extreme low-light and sparse-view scenarios in the NTIRE 2026 3D Restoration and Reconstruction Challenge \cite{liu2025realx3d}. ELoG-GS follows a restoration-then-reconstruction framework to address issues such as geometric collapse, color drift, and noise-induced failures commonly found in conventional SfM and vanilla 3DGS methods \cite{schonberger2016structure,kerbl20233d}. By employing VGGT depth estimation and voxelized fusion for reliable point cloud initialization \cite{wang2025vggt}, Retinexformer for zero-shot low-light enhancement \cite{cai2023retinexformer}, and a dual-branch architecture combining FSGS and EAP-GS for stable high-fidelity reconstruction \cite{zhu2024fsgs,dai2025eap}, our method achieves strong geometric consistency and photorealistic rendering. Luminance-guided post-processing further improves color fidelity and eliminates artifacts.
Experiments on the challenge benchmark demonstrate that ELoG-GS outperforms baseline methods in both quantitative metrics and visual quality, offering an effective solution for real-world low-light 3D reconstruction. Future work will focus on end-to-end joint optimization, automatic branch selection, and the integration of more expressive splatting primitives to further improve reconstruction fidelity in degraded scenarios.

% \section*{Acknowledgments}
% Any acknowledgments or funding information.

{
    \small
    \bibliographystyle{ieeenat_fullname}
    \bibliography{main}
}

% \bibliography{references}
\end{document}